\documentclass[10pt,twocolumn]{article}

\usepackage[T1]{fontenc}
\usepackage[utf8]{inputenc}
\usepackage{lmodern}
\usepackage{microtype}
\usepackage[hidelinks]{hyperref}
\usepackage[margin=1in]{geometry}

\usepackage{xcolor}
\usepackage{amsmath,amssymb}
\usepackage{placeins}
\usepackage{abstract}
\usepackage{float}
\usepackage{caption}
\usepackage{booktabs}
\usepackage{multicol}
\usepackage{array}
\usepackage{graphicx}
\usepackage{tabularx}
\usepackage{longtable}
\usepackage{calc}
\usepackage{etoolbox}
\usepackage{bookmark}
\usepackage{soul}
\usepackage{fancyhdr}

\def\fps@figure{htbp}

\IfFileExists{footnotehyper.sty}{\usepackage{footnotehyper}}{\usepackage{footnote}}
\makesavenoteenv{longtable}

\usepackage{titlesec}
\titleformat{\section}
  {\normalfont\large\bfseries\raggedright}{\thesection}{0.75em}{}
\titlespacing*{\section}{0pt}{2.0ex plus .4ex}{0.8ex}

\titleformat{\subsection}
  {\normalfont\normalsize\bfseries\raggedright}{\thesubsection}{0.75em}{}
\titlespacing*{\subsection}{0pt}{1.5ex plus .3ex}{0.6ex}

\titleformat{\subsubsection}
  {\normalfont\normalsize\itshape\raggedright}{\thesubsubsection}{0.75em}{}
\titlespacing*{\subsubsection}{0pt}{0.8ex plus .2ex}{0.4ex}

\fancypagestyle{firstpage}{%
  \fancyhf{} 
  
  \fancyfoot[C]{\small *~Corresponding authors:
    \href{mailto:pratnaik@paypal.com}{pratnaik@paypal.com},
    \href{mailto:ndintakurthi@paypal.com}{ndintakurthi@paypal.com},
    \href{mailto:zhanhu@paypal.com}{zhanhu@paypal.com},
    \href{mailto:ywang49@paypal.com}{ywang49@paypal.com},
    \href{mailto:xqiu1@paypal.com}{xqiu1@paypal.com}}
}

\title{\textbf{Co-Investigator AI: The Rise of Agentic AI for Smarter, Trustworthy
AML Compliance Narratives}}
\author{Prathamesh Vasudeo Naik$^{*}$ \and Naresh Kumar Dintakurthi$^{*}$ \and Zhanghao Hu$^{*}$ \and Yue Wang$^{*}$ \and Robby Qiu$^{*}$}
\date{}

\begin{document}
\renewcommand{\abstractname}{\large\bfseries Abstract}

\twocolumn[
  \begin{@twocolumnfalse}
    \vspace*{-2cm}
    \maketitle
    \thispagestyle{firstpage} 
    \begin{abstract}
    \vspace{2em}
    Generating regulatorily compliant Suspicious Activity Report (SAR)
    remains a high-cost, low-scalability bottleneck in Anti-Money Laundering
    (AML) workflows. While large language models (LLMs) offer promising
    fluency, they suffer from factual hallucination, limited crime typology
    alignment, and poor explainability---posing unacceptable risks in
    compliance-critical domains. This paper introduces
    \textbf{Co-Investigator AI}, an agentic framework optimized to produce
    Suspicious Activity Reports (SARs) significantly faster and with greater
    accuracy than traditional methods. Drawing inspiration from recent
    advances in autonomous agent architectures, such as the AI
    Co-Scientist\hyperlink{ref-2}{[2]}, our approach integrates specialized agents for
    planning, crime type detection, external intelligence gathering, and
    compliance validation. The system features dynamic memory management, an
    AI-Privacy Guard layer for sensitive data handling, and a real-time
    validation agent employing the Agent-as-a-Judge paradigm\hyperlink{ref-5}{[5]} to
    ensure continuous narrative quality assurance. Human investigators
    remain firmly in the loop, empowered to review and refine drafts in a
    collaborative workflow that blends AI efficiency with domain expertise.
    We demonstrate the versatility of Co-Investigator AI across a range of
    complex financial crime scenarios, highlighting its ability to
    streamline SAR drafting, align narratives with regulatory expectations,
    and enable compliance teams to focus on higher-order analytical work.
    This approach marks the beginning of a new era in compliance
    reporting---bringing the transformative benefits of AI agents to the
    core of regulatory processes and paving the way for scalable, reliable,
    and transparent SAR generation.

    \end{abstract}
    \vspace{2em}
  \end{@twocolumnfalse}
]

\section{Introduction}\label{introduction}

Compliance teams at financial institutions (FIs) routinely produce SAR
narratives---crucial documents that elucidate suspicious conduct for
regulatory and law-enforcement review. However, crafting these
narratives is time-consuming and inconsistent. FinCEN estimates the
final stages of SAR preparation take between \textbf{25 and 315
minutes}, depending on complexity. This delay results from manually
gathering transactional data, KYC records, communication metadata, and
investigative context. Moreover, narrative quality varies widely across
FIs, driven by differences in investigator experience, workload, and
interpretive judgment---introducing compliance risks. Concurrently,
mounting transaction volumes and evolving crime typologies (e.g.,
cryptocurrency layering, elder exploitation, digital fraud) are
overwhelming traditional workflows.

These challenges underscore an urgent need for scalable, reliable
systems that ensure narrative quality and regulatory compliance while
alleviating investigator burden. This paper presents
\textbf{Co‑Investigator AI}---a human-in-the-loop, agentic system that
automates data parsing, crime type detection, and draft narrative
generation, empowering FIs to enhance both investigator productivity and
compliance robustness.

\section{AML, SAR and Compliance Landscape}\label{aml-sar-and-compliance-landscape}

The \textbf{Suspicious Activity Report (SAR)} is a cornerstone in
Anti-Money Laundering (AML) compliance, acting as the primary mechanism
for financial institutions (FIs) to report suspicious activities---such
as fraud, money laundering, terrorist financing, and sanctions
evasion---to regulators like FinCEN in the U.S. SARs were
institutionalized under the \textbf{Bank Secrecy Act of 1970}, with the
report format standardized in 1996 to ensure consistent and effective
reporting practices. Since July 1, 2012, U.S. SARs are filed
electronically, improving timeliness and data integrity.

\subsection{Regulatory Requirements and SAR Content}\label{req-sar}

U.S. financial institutions must electronically file SARs within
\textbf{30 calendar days} from the initial detection of suspicious
activity. If the subject is unidentified, the reporting window may be
extended to \textbf{60 days} in total. In cases of extreme
urgency---such as suspected terrorist financing---immediate filings and
direct notification of law enforcement are mandatory.

FinCEN requires SARs to include:

\begin{enumerate}
\def\labelenumi{\arabic{enumi}.}
\item
  \textbf{Subject Details}: Identifying information (name, address, date
  of birth, SSN, associated accounts).
\item
  \textbf{Suspicious Activity Description}: A date range and
  classification code describing suspicious behavior.
\item
  \textbf{Institution Information}: Details about the reporting FI and
  the relevant branch, platform, or system.
\item
  \textbf{Filer Contact Details}: Compliance officer's full contact info
  for regulatory follow-up.
\item
  \textbf{Narrative Explanation}: A comprehensive account covering who,
  what, when, where, and how the activity unfolded.
\item
  \textbf{Supporting Documentation}: Transaction logs, communications,
  account records retained for \textbf{at least five years}.
\end{enumerate}

\subsection{Challenges in Narrative Generation}\label{challenge-nar}

The \textbf{narrative section} is both the most critical and the most
resource-intensive component of the SAR. It must convey a clear, legally
compliant story without ``tipping off'' the subject---making it a
complex blend of factual clarity and strategic discretion. Traditional
methods---manual drafting, disparate tool usage, and data
aggregation---are time-consuming and subject to interpretation,
resulting in potential errors, inconsistencies, and compliance
vulnerabilities.

\subsection{Compliance Pressure and Institutional Burden}\label{compliance-burden}

FIs face increasingly stringent regulatory scrutiny from bodies such as
FinCEN, the Financial Action Task Force (FATF), and other global
regulators. Non-compliance---including late SAR filings or incomplete
narratives---can trigger severe penalties, reputational harm, and
heightened supervisory oversight.

Furthermore, SARs serve a dual role: they aid immediate law enforcement
investigations and contribute to long-term trend analysis and regulatory
strategy development based on aggregated SAR data.

\subsection{Relevance to Our Research}\label{relevance-research}

Given the escalating data volumes, increasing transaction complexity,
and stricter AML regulatory demands, financial institutions urgently
need scalable, accurate, and explainable approaches to SAR generation.
Our \textbf{Co‑Investigator AI} addresses these issues by:

\begin{itemize}
\item
  Automating narrative drafting while maintaining human
  interpretability.
\item
  Ensuring narratives meet structure and substance requirements mandated
  by regulators.
\item
  Accelerating investigation timelines without sacrificing compliance
  quality.
\end{itemize}

By aligning with established SAR frameworks---timeliness requirements,
content mandates, and narrative best practices---our research aims to
strike the balance between accelerated workflow and rigorous regulatory
adherence.

\section{Why Traditional SAR Workflows Break: Data Friction, Scalability Limits, and GenAI Gaps}\label{why-traditional-sar-workflows-break-data-friction-scalability-limits-and-genai-gaps}

Financial institutions continue to rely heavily on manual workflows to
generate Suspicious Activity Reports (SARs), despite the growing
complexity and volume of financial crimes. These workflows---while
human-guided and context-rich---are increasingly inadequate under the
pressures of modern AML operations. Institutions face tighter regulatory
timelines, growing typology diversity, and rising expectations around
auditability and consistency.

While recent advances in large language models (LLMs) have opened the
door to semi-automated text generation, the limitations of unstructured,
monolithic GenAI applications remain poorly understood in compliance
contexts. Before proposing a modular agentic alternative, we first
characterize the failure modes of traditional and baseline LLM-driven
SAR generation methods.

\subsection{Manual Drafting: Fragmented Tools, High Latency, and Cognitive Overload}\label{manual-drafting-fragmented-tools-high-latency-and-cognitive-overload}

Suspicious Activity Reports are synthesized from multiple sources:
structured transaction data, account metadata, communication logs,
external alerts, and KYC profiles. Investigators manually gather and
interpret this information across siloed tools like spreadsheets, case
management dashboards, and third-party research platforms.

This workflow presents multiple systemic limitations:

\begin{itemize}
\item
  \textbf{Complexity \& Time Burden}: Investigators often spend several
  hours per narrative due to the intricate nature of the data and
  required regulatory detail.
\item
  \textbf{Inconsistency}: Narrative quality is susceptible to variance
  in experience and style, negatively affecting interpretability.
\item
  \textbf{Scalability Limits}: Increasing SAR volumes overwhelm manual
  capacities, risking incomplete or delayed filings.
\item
  \textbf{Error Risk}: Human-curated narratives are prone to omissions,
  misinterpretation, and potential regulatory non-compliance.
\item
  \textbf{Operational Overhead}: Investigators frequently juggle
  multiple tools---Excel, case management systems, external
  databases---leading to fragmented workflows and reduced efficiency.
\end{itemize}

These constraints underscore the pressing need for automation
technologies that maintain narrative rigor while reducing
investigators\textquotesingle{} cognitive and operational burdens.

\subsection{Limitations of Direct GenAI Narrative Generation}\label{limitations-of-direct-genai-narrative-generation}

Recent research highlights the increasing adoption of AI techniques in
AML contexts\hyperlink{ref-4}{[4]}\hyperlink{ref-6}{[6]}. To explore whether modern LLMs can bypass
these bottlenecks, we tested direct prompting approaches using raw AML
case data. While LLMs (e.g., GPT-4, Claude 3) performed reasonably on
clean, low-complexity cases, results degraded sharply in real-world
conditions.

\begin{itemize}
\item
  \textbf{Template Scenarios}: For highly structured, low-complexity
  cases, the model produced coherent drafts that aligned with expected
  formats, showing promise in straightforward contexts.
\item
  \textbf{Performance Breakdown}: Once confronted with real-world
  complexity---multiple subjects, overlapping financial crime
  typologies, and extensive transaction histories---the model produced
  inconsistent and unreliable outputs.
\item
  \textbf{Hallucination Risks}: The LLM invented details such as
  fabricated interactions, spurious transaction events, and contextually
  unsupported notes, even when explicitly instructed not to add
  information. This aligns with documented hallucination
  rates---frequently exceeding 20--30\% in LLM-generated content\hyperlink{ref-7}{[7]}.
\item
  \textbf{Review Cost}: Although drafting appeared rapid, the required
  manual validation and correction nullified any time benefits, shifting
  the burden to investigators.
\end{itemize}

These limitations highlight a critical gap: LLMs can produce
grammatically correct outputs, but lack \textbf{traceability, modular
reasoning, and typology-specific sensitivity}. This motivates a shift
away from monolithic generation toward \textbf{agentic decomposition},
where specialized AI agents reason, validate, and interact in concert
with human investigators.

\section{Agentic AI and the Co Investigator Paradigm: Enhancing SAR Generation}\label{agentic-ai-and-the-co-investigator-paradigm-enhancing-sar-generation}

To address the shortcomings revealed by direct Gen AI experimentation,
we propose the \textbf{Co‑Investigator Paradigm}---a fully agentic AI
framework tailored for high-stakes compliance tasks. This paradigm draws
inspiration from the multi-agent coordination seen in systems like the
AI Co‑Scientist\hyperlink{ref-2}{[2]} but adapts these concepts for financial crime
investigations.

\begin{figure}[htbp]
  \centering
  \includegraphics[width=\linewidth]{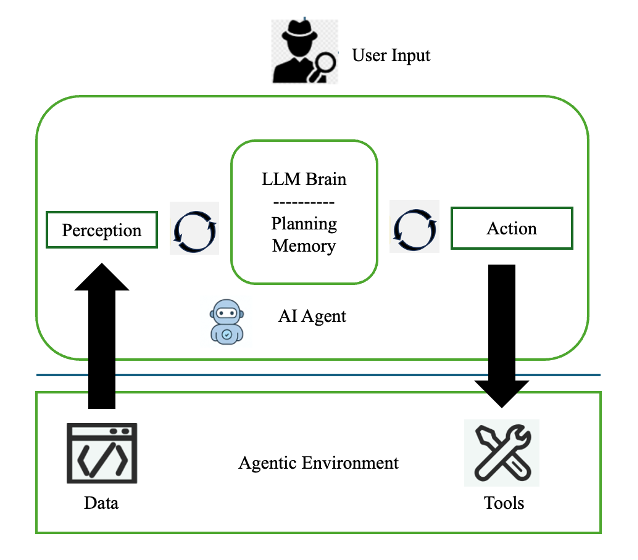}
  \caption{Agentic AI in action: Perceive--Reason--Act architecture}
  \label{fig:modular-architecture}
\end{figure}

\textbf{Agentic AI}, as surveyed by Xi et al\hyperlink{ref-9}{[9]}, defines intelligent
systems that can perceive inputs, reason adaptively, and act\hyperlink{ref-23}{[23]}
---often collaboratively in multi-agent settings or alongside
humans\hyperlink{ref-10}{[10]}. AI systems are becoming increasingly capable---a trend
that is likely to accelerate with the development of more agentic AI
systems in the future\hyperlink{ref-14}{[14]}\hyperlink{ref-15}{[15]}. Co‑Investigator AI embodies these
principles, featuring purpose-built agents that independently:

\begin{enumerate}
\def\labelenumi{\arabic{enumi}.}
\item
  \textbf{Perceive}: ingest and normalize transaction data, KYC records,
  and external intelligence.
\item
  \textbf{Reason}: identify typologies, assess regulatory compliance,
  and draft coherent narratives using transparent Chain-of-Thought
  prompting\hyperlink{ref-3}{[3]}.
\item
  \textbf{Act}: generate initial SAR drafts and flag compliance concerns
  in real-time.
\end{enumerate}

This agentic structure supports a \textbf{human‑agent collaborative
ecosystem}. Specialized agents---responsible for tasks such as crime
type detection, narrative drafting, and compliance validation---work in
concert under human oversight. Investigators receive structured SAR
drafts and refine them using iterative feedback guided by the model.
This \textbf{human-in-the-loop} design maintains accountability and
interpretability, ensuring outputs meet regulatory standards and
cultural nuances.

By combining autonomous agentic operations with human expertise,
Co‑Investigator AI delivers scalable, transparent, and reliable SAR
narrative generation---far surpassing monolithic LLM approaches in both
performance and auditability.

\section{Detailed System Design}\label{detailed-system-design}

The \textbf{Co‑Investigator AI} deploys a modular, agentic architecture (Figure 2)
for purpose‑built efficient, high‑quality Suspicious Activity Report
(SAR) narrative generation. The system orchestrates specialized AI
agents supported by dynamic memory, investigator feedback loops,
automated compliance checks, and explicit reasoning protocols. This
cohesive, human‑centered framework significantly improves scalability,
narrative accuracy, and adherence to regulatory standards.

\subsection{Agentic Modular Architecture Overview}\label{agentic-modular-architecture-overview}

The Co-Investigator AI is structured as a sophisticated, modular,
agentic architecture designed explicitly for the automated generation
and iterative refinement of Suspicious Activity Report (SAR) narratives.
This architecture encompasses distinct yet interconnected functional
agents, ensuring precise compliance analysis, efficient workflow
orchestration, transparent reasoning processes, and structured
human-agent collaboration. Key modules within the system include:

\begin{itemize}
\item Data Ingestion And Structuring Layer 
\item AI-Privacy Guard Layer
\item Crime Type Detection
\item Planning Agent
\item Typology Detection Agents
\item External Intelligence Agent
\item Narrative Generation Agent
\item Compliance Validation Agent
\item Feedback Agent
\item Dynamic Memory
\item Analytical Tools
\end{itemize}

\begin{figure*}[!ht]
  \centering
  \includegraphics[width=\textwidth]{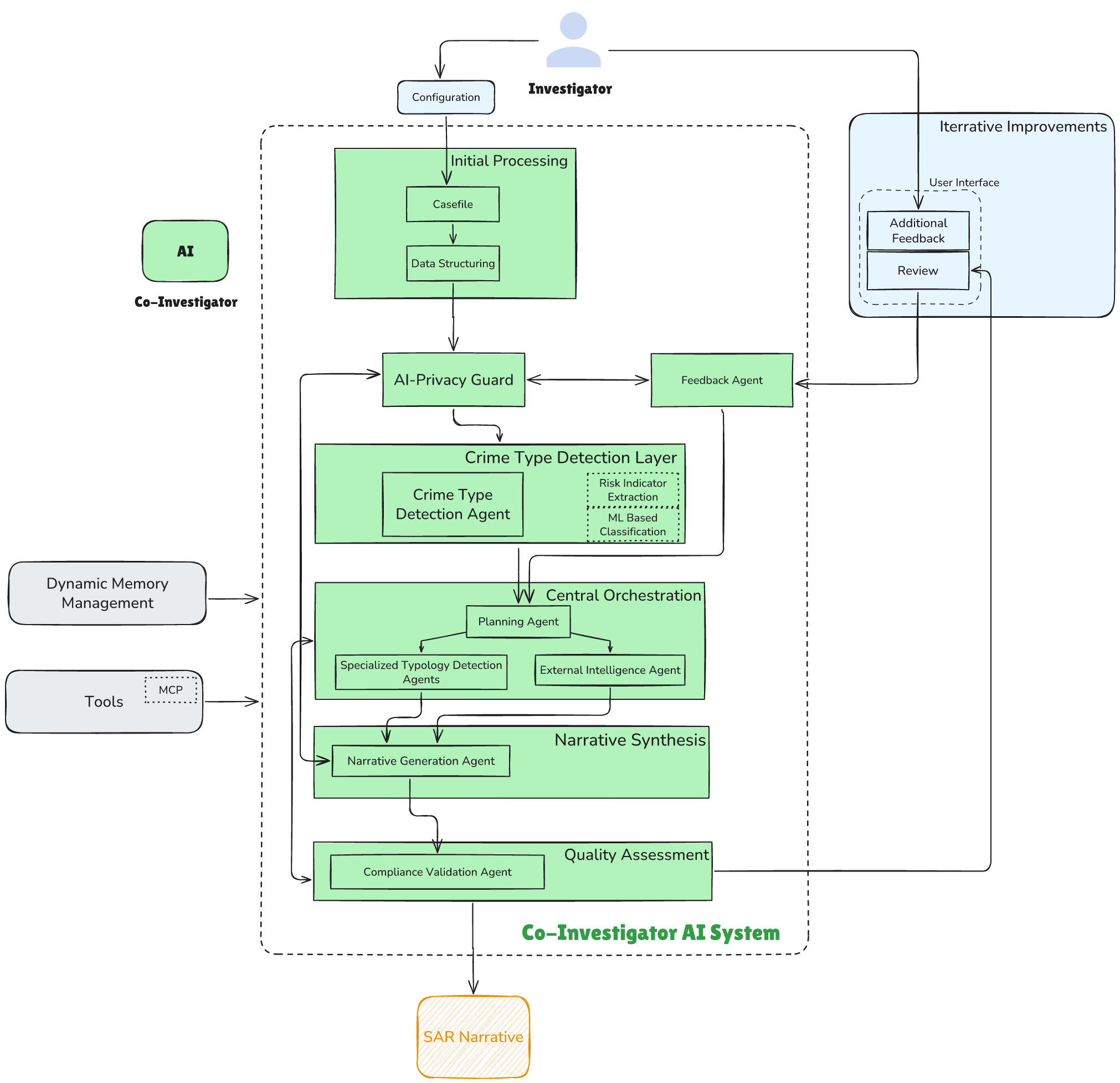}
  \caption{Modular agentic architecture of Co-Investigator AI for SAR generation}
  \label{fig:modular-architecture}
\end{figure*}

\subsection{Data Ingestion \& Structuring Layer}\label{data-ingestion-structuring-layer}

The process begins with the \textbf{Data Ingestion \& Structuring
Layer}, which operates independently of the central orchestration
component. This layer systematically ingests raw alert
data---transaction records, account metadata, customer identifiers, and
risk signals---and transforms it into structured summaries. These parsed
outputs provide the foundational inputs for the downstream
\textbf{AI‑Privacy Guard} and subsequent analytic pipeline.

\subsection{AI-Privacy Guard Layer}\label{ai-privacy-guard-layer}

To address critical data privacy and compliance requirements, our
architecture integrates with a dedicated AI-Privacy Guard layer
positioned upstream of the Large Language Model (LLM) processing
pipeline. The primary objective of this layer is to systematically
identify and anonymize sensitive entities, specifically data categorized
as class-1 or class-2 confidential information, prior to submission to
external LLM services. This AI-Privacy Guard layer serves as a
horizontal capability ensuring robust protection and regulatory
compliance of sensitive data.

Unlike conventional Named Entity Recognition (NER) tasks, our model
faces distinct challenges due to the nature and variability of input
data:

\begin{itemize}
\item
  \textbf{Highly Unstructured Input Data:} Input data originates from
  diverse formats including source code (HTML, etc.), conversational
  text, and miscellaneous free-form prompts, creating substantial
  variability and complexity.
\item
  \textbf{Extended Context Length:} Inputs frequently reach the maximum
  token limits of contemporary LLMs, demanding models capable of
  efficiently processing lengthy, contextually dense texts without loss
  of performance or accuracy.
\item
  \textbf{Critical SLA Requirements:} Real-time performance and strict
  adherence to service-level agreements (SLAs) are mandatory, requiring
  rapid inference even on extensive, highly unstructured text inputs
  containing potentially unlimited sensitive entities.
\end{itemize}

To effectively address these specific demands, our AI-Privacy Guard
solution employs a robust \textbf{RoBERTa\hyperlink{ref-16}{[16]} model combined with}
\textbf{Conditional Random Fields (CRF)}, optimized through advanced
methods including model quantization and deep speed-based inference
acceleration. This optimization approach ensures both computational
efficiency and consistent compliance with stringent SLAs, while
maintaining high-precision anonymization capabilities across a diverse
array of complex, lengthy, and sensitive data inputs.

AI-Privacy Guard layer operates dynamically across multiple agents,
facilitating secure interactions at critical workflow stages:

\begin{itemize}
\item
  \textbf{Pre-Processing Integration:} After the Data Ingestion and
  Structuring Layer processes raw data, the AI-Privacy Guard anonymizes
  sensitive elements before data is routed to crime type detection
  agents and external LLMs.
\item
  \textbf{Typology Agents Integration:} Ensures sensitive data is
  consistently masked across all specialized typology detection agents
  during their analytical processing, maintaining robust data protection
  and compliance.
\item
  \textbf{Narrative Generation Interaction:} After analytical
  processing, the anonymized data is unmasked by the AI-Privacy Guard
  before incorporation into the narrative drafts, enabling investigators
  to interpret and validate precise, contextually rich narratives.
\item
  \textbf{Feedback Agent Integration:} Prior to transmitting narratives
  to external feedback systems or review interfaces, the AI-Privacy
  Guard re-masks sensitive details to ensure continued confidentiality
  during human-in-the-loop interactions.
\end{itemize}

\subsection{Crime Type Detection Layer}\label{crime-type-detection-layer}

Accurate crime type detection is a critical yet resource-intensive
component within AML compliance and Suspicious Activity Report (SAR)
generation. Financial crime investigations frequently involve multiple,
overlapping types, such as elder financial exploitation, romance scams,
human trafficking, money mule schemes, terrorist financing, CSAM,
identity theft, and various fraud schemes. Traditional manual
classification approaches consume substantial investigative resources,
underscoring the need for systematic automation.

The Co-Investigator AI addresses this challenge through a specialized
\textbf{Crime Type Detection Agent}, designed with two integrated
analytical components:

\begin{itemize}
\item
  \textbf{Risk-Indicator Extraction Tools:}
  The Crime Type Detection Agent leverages automated risk-based tools to
  scan all relevant compliance data---both structured records and
  free-form text---and surface key risk indicators. It abstracts away
  low-level details, instead flagging broad ``red flag'' categories
  (e.g., anomalous access, atypical payment behavior, and potential
  exploitation patterns) for downstream analysis.
\item
  \textbf{Machine Learning Classification Models:}
  Complementing these automated tools, sophisticated tree-based ensemble
  models (e.g., Random Forest, Gradient Boosting) systematically analyze
  extracted risk indicators alongside historical data, transaction
  patterns, and entity relationships. These models produce probabilistic
  assessments of potential financial crime typologies, allowing precise
  multi-typology classification.
\end{itemize}

The combined outputs---ranked typologies with associated confidence
scores---inform the Central Orchestration and Planning Agent, which
dynamically coordinates specialized analytical agents tailored to each
identified type. This systematic, integrated approach significantly
reduces investigator workloads, enhances narrative accuracy, and ensures
rigorous compliance, underscoring the operational effectiveness of
automated typology detection.

\subsection{Planning Agent}\label{planning-agent}

The \textbf{Planning Agent} serves as the central orchestrator within
the agentic architecture, primarily responsible for dynamically spawning
specialized typology detection agents and external intelligence agents.
It leverages outputs from the crime type detection agent, real-time
feedback from operational agents, and structured investigator input to
determine which agents to activate. This dynamic orchestration ensures
precise resource allocation, optimized investigative efficiency, and
consistent regulatory compliance throughout the SAR generation
lifecycle.

\subsection{Specialized Typology Detection Agents}\label{specialized-typology-detection-agents}

The following specialized agents systematically analyze specific
financial crime typologies, each dedicated to distinct risk-detection
tasks:

\begin{itemize}
\item
  \textbf{Transaction Fraud Detection Agent:} Detects suspicious
  transaction patterns indicative of potential fraud.
\item
  \textbf{Payment Volume Velocity Detection Agent:} Flags abnormal
  transaction frequencies or high-volume activity.
\item
  \textbf{Country Risk Detection Agent:} Evaluates transactions linked
  to high-risk or sanctioned jurisdictions.
\item
  \textbf{Textual Content Detection Agent:} Identifies anomalies in
  textual customer communications and transaction notes using advanced
  NLP.
\item
  \textbf{Geographic Anomaly Detection Agent:} Detects location-based
  inconsistencies and transaction anomalies.
\item
  \textbf{Account Health Assessment Agent:} Assesses historical account
  activities for potential compromise or misuse.
\item
  \textbf{Dispute Pattern Detection Agent:} Analyzes dispute patterns to
  detect suspicious or fraudulent claims.
\end{itemize}

\subsection{External Intelligence Agent (with MCP Integration)}\label{external-intelligence-agent-with-mcp-integration}

To enrich SAR narratives with pertinent external risk intelligence, our
system incorporates an \textbf{External Intelligence Agent} that
leverages the \textbf{Model Context Protocol (MCP)\hyperlink{ref-24}{[24]}} an emerging
open standard for secure, tool-agnostic AI integration. MCP enables
seamless and dynamic access to external data sources (e.g., news feeds,
sanctions list, media reports) through standardized servers without
requiring bespoke API integrations.

\paragraph{\texorpdfstring{\textbf{Key Design
Features}}{Key Design Features}}\label{key-design-features}

\begin{itemize}
\item
  \textbf{Standardized Tool Discovery \& Invocation}\\
  The agent dynamically identifies data sources exposed via MCP
  servers---treating each as a tool with well-defined metadata. This
  allows it to select and invoke sources at runtime without
  personalization or hard-coded logic.
\item
  \textbf{Context-Driven Intelligence Retrieval}\\
  Upon crime type detection or investigator request (e.g., linking a
  subject to a high-risk jurisdiction), the agent issues MCP queries.
  Returned data---such as negative news articles, regulatory advisories,
  or watchlist flags---is integrated into narrative drafts, providing
  timely contextualization.
\end{itemize}

\subsection{Narrative Generation Agent}\label{narrative-generation-agent}

Integrating analytical insights from previous modules, the
\textbf{Narrative Generation Agent} synthesizes structured SAR narrative
drafts. Utilizing explicit Chain-of-Thought (CoT)\hyperlink{ref-3}{[3]}\hyperlink{ref-30}{[30]}
prompting methods, it transparently articulates its reasoning,
aggregating risk indicators, transaction details, external intelligence,
and historical regulatory context. This agent generates comprehensive
and structured drafts explicitly intended for investigator review and
refinement.

\begin{figure}[htbp]
  \centering
  \includegraphics[width=\linewidth]{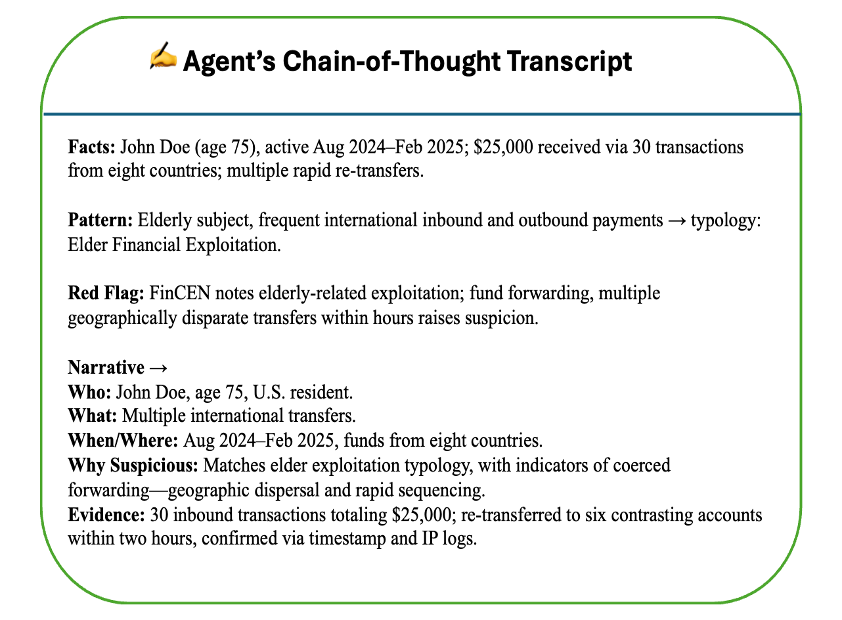}
  \caption{Chain-of-Thought reasoning within Co-Investigator AI's
narrative generation}
  \label{fig:chain-of-thought}
\end{figure}

\subsection{Compliance Validation and Explicit Reasoning}\label{compliance-validation-and-explicit-reasoning}

\begin{itemize}
\item
  \textbf{Compliance Validation Agent (Agent-as-a-Judge):}
  Employs rigorous automated evaluation methodologies---combining
  semantic coherence assessments and rule-based regulatory accuracy
  checks---to systematically verify narrative quality and compliance
  alignment. Guided explicitly by the Agent-as-a-Judge
  methodology\hyperlink{ref-5}{[5]}, this agent ensures consistently high-quality SAR
  narratives.
\end{itemize}

\begin{itemize}
\item
  \textbf{Confidence Score-Based Reasoning Framework:}
  Within the Chain-of-Thought methodology, agents assign structured
  confidence scores reflecting evidentiary strength, contextual
  relevance, and regulatory adherence. This transparent scoring
  mechanism ensures clear interpretability, accountability, and
  investigator trust in the generated narrative components. These
  techniques are also used to estimate LLM confidence in the factual
  accuracy of its responses\hyperlink{ref-32}{[32]}.
\end{itemize}

\subsection{Feedback Agent for Human-in-the-Loop Refinement}\label{feedback-agent-for-human-in-the-loop-refinement}

The dedicated \textbf{Feedback Agent} systematically integrates
structured human feedback into iterative refinement processes.
Investigators provide narrative reviews and adjustments through a secure
editing interface, with their inputs explicitly captured and dynamically
incorporated into subsequent narrative cycles, ensuring continuous
improvements and refined narrative precision. This mirrors recent work
on feedback-driven refinement frameworks such as \emph{Self‑Refine},
which employs iterative feedback loops to significantly enhance LLM
output quality without requiring additional training data\hyperlink{ref-33}{[33]}.

\subsection{Dynamic Memory Management}\label{dynamic-memory-management}

To enhance narrative coherence and regulatory consistency, the
architecture employs sophisticated memory management techniques across
multiple layers:

\begin{itemize}
\item
  \textbf{Regulatory Memory:} Maintains up-to-date regulatory
  guidelines, AML standards, and compliance expectations, ensuring
  ongoing regulatory compliance.
\item
  \textbf{Historical Narrative Memory:} Stores prior SAR narratives and
  investigative outcomes, promoting consistent narrative generation and
  continuity.
\item
  \textbf{Typology-Specific Memory:} Captures and recalls historical
  typology-specific analytical patterns, risk indicators, and prior
  investigative insights, enabling contextually precise risk assessment
  and narrative accuracy.
\end{itemize}

While \textbf{Retrieval-Augmented Generation (RAG)} pipelines\hyperlink{ref-37}{[37]}
typically inject relevant documents into prompt contexts at inference
time, they are limited by stateless execution and prompt length
constraints. In contrast, Co-Investigator AI\textquotesingle s memory
system provides \textbf{persistent and updateable memory layers},
tightly integrated into agent workflows and continuously synchronized
across tasks.

This multi-tiered memory strategy reflects recent advancements in
hierarchical and agentic memory systems for LLM agents. For example,
\textbf{MemoryOS\hyperlink{ref-34}{[34]}} introduces a three-level memory
hierarchy---short-term, mid-term, and long-term---and dynamic update
mechanisms to preserve contextual integrity over extended interactions.
Similarly, \textbf{A-MEM\hyperlink{ref-35}{[35]}} implements an agentic, structured
memory network inspired by Zettelkasten methods, dynamically indexing
and evolving memory content based on contextual relevance. Together,
these frameworks underline the importance of structured memory design
for maintaining system coherence, recall accuracy, and longitudinal
consistency in agentic workflows.

\subsection{Supporting Analytical Tools}\label{supporting-analytical-tools}

To further enhance the analytical capabilities of our agentic system, we
have integrated specialized analytical tools designed to accelerate and
deepen investigative processes. These tools systematically analyze data,
extract critical insights, and enrich investigative contexts, thereby
significantly improving the accuracy, efficiency, and interpretability
of compliance investigations. Specifically, the supporting analytical
tools include:

\begin{itemize}
\item
  \textbf{Risk Indicator Extraction Tool:}
  Automatically scans structured and unstructured data to proactively
  identify and extract critical risk indicators, streamlining subsequent
  investigative efforts and enhancing narrative precision.
\end{itemize}

\begin{itemize}
\item
  \textbf{External Intelligence Search Tool:}
  Integrates external data sources, such as negative news, sanctions
  alerts, and regulatory advisories, via secure search APIs. This tool
  provides relevant external context to investigations, significantly
  enriching the interpretative depth of narratives.
\item
  \textbf{Account-Linking Analysis Tool:}
  Systematically identifies hidden relationships and interconnected
  patterns among accounts, entities, and subjects. By highlighting
  complex networks of suspicious activity, it supports investigators in
  uncovering organized financial crime schemes.
\end{itemize}

Collectively, these analytical tools serve as essential support
mechanisms for our specialized typology detection agents. By proactively
extracting relevant risk indicators, seamlessly integrating external
intelligence, and uncovering complex relational patterns, these tools
significantly enhance investigative depth, efficiency, and accuracy. As
a result, the Co-Investigator AI system consistently generates detailed,
compliant, and comprehensive SAR narratives, greatly augmenting the
effectiveness of compliance investigations.
\section{Agentic Architecture: Sub-Agent Pattern}\label{agentic-architecture-subagent-pattern}

In designing Co-Investigator AI system, we adopted a sub-agent
architecture, inspired by modular decomposition principles analogous to
microservices. This design decomposes a complex compliance workflow into
independently operating, specialized agents---each aligned to a distinct
function (e.g., crime type detection, narrative generation, compliance
validation). This approach enhances system robustness, adaptability, and
maintainability.

Benefits of the Sub-Agent Pattern:

\begin{enumerate}
\def\labelenumi{\arabic{enumi}.}
\item
  \textbf{Isolation and Fault Tolerance}: Each sub-agent is an
  encapsulated module responsible for a focused task. Failures or
  updates in a single agent---such as the Narrative Generation
  Agent---do not cascade across the system. This reduces system-wide
  downtime and simplifies error diagnostics, enhancing overall fault
  tolerance.
\item
  \textbf{Scalability and Maintainability}: Agents can be scaled,
  optimized, or replaced individually without disrupting other
  components. This modularity allows for focused updates, enabling
  faster development cycles and easier adaptation to evolve compliance
  requirements or new typologies.
\item
  \textbf{Dynamic Orchestration}: The Planning Agent serves as a
  lightweight orchestrator, dynamically spawning sub-agents based on
  crime type detection results or investigator feedback. Unlike
  monolithic, code-heavy routing logic, this event-driven approach
  enables flexibility and cleaner workflows---agents operate
  independently yet collaborate cohesively.
\item
  \textbf{Frictionless Innovation}: The modular design encourages
  targeted experimentation. Enhancing one agent---for example,
  integrating a new ML model into the Crime Type Detection Agent---can
  be done independently. This promotes clean feature rollouts, quicker
  experimentation, and easier regression control across modules.
\end{enumerate}

By adopting a sub-agent design, our framework achieves clarity,
adaptability, and resilience---empowering domain experts to scale,
evolve, and maintain the system efficiently in response to changing
regulatory landscapes and investigative demands.

\section{Automated Evaluation Framework for Rapid Iteration}\label{automated-evaluation-framework-for-rapid-iteration}

To systematically validate the performance of the Co-Investigator AI
framework and ensure the consistent generation of high-quality
Suspicious Activity Report (SAR) narratives, we established an automated
pre-production evaluation framework (Figure 4) specifically designed for
accelerated iterative development. Developed collaboratively with the
compliance investigators, this evaluation system leverages expertly
annotated "golden datasets" as authoritative benchmarks, enabling rapid
assessment, transparent scoring, and continuous model refinement prior
to deployment in live operational environments. Key elements of the
pre-production evaluation include:

\textbf{Objective Comparative Evaluation:}

\begin{itemize}
\item
  Systematically compares AI-generated narratives against
  investigator-crafted SAR narratives.
\item
  Employs a hybrid evaluation approach combining rule-based logical
  assessments and semantic similarity analyses powered by Large Language
  Models (LLMs).
\end{itemize}

\textbf{Structured Multi-Component Scoring:}

\begin{itemize}
\item
  \textbf{Intro Scoring:}
  Precisely evaluates narrative introductory elements---such as date
  ranges, transaction amounts, and subject identification---using
  weighted accuracy checks to ensure factual correctness and regulatory
  compliance.
\item
  \textbf{Narrative Scoring:}
  Assesses the narrative body across seven distinct compliance
  dimensions, applying semantic similarity metrics with configurable
  category-specific weights to rigorously measure coherence, analytical
  completeness, and regulatory adherence.
\end{itemize}

\textbf{Aggregated Weighted Final Scoring:}

\begin{itemize}
\item
  Combines the Intro and Narrative scoring components into a single,
  transparent weighted final score.
\item
  Enables quantitative benchmarking against regulatory standards,
  promoting clear interpretability and structured iterative improvements
  to SAR narrative outputs.
\end{itemize}

\textbf{Early Performance Validation:}
  Demonstrates consistently high-quality scores effectively evaluating
  complex financial crime typologies such as elderly financial
  exploitation.

\begin{figure}
  \centering
  \includegraphics[width=\linewidth]{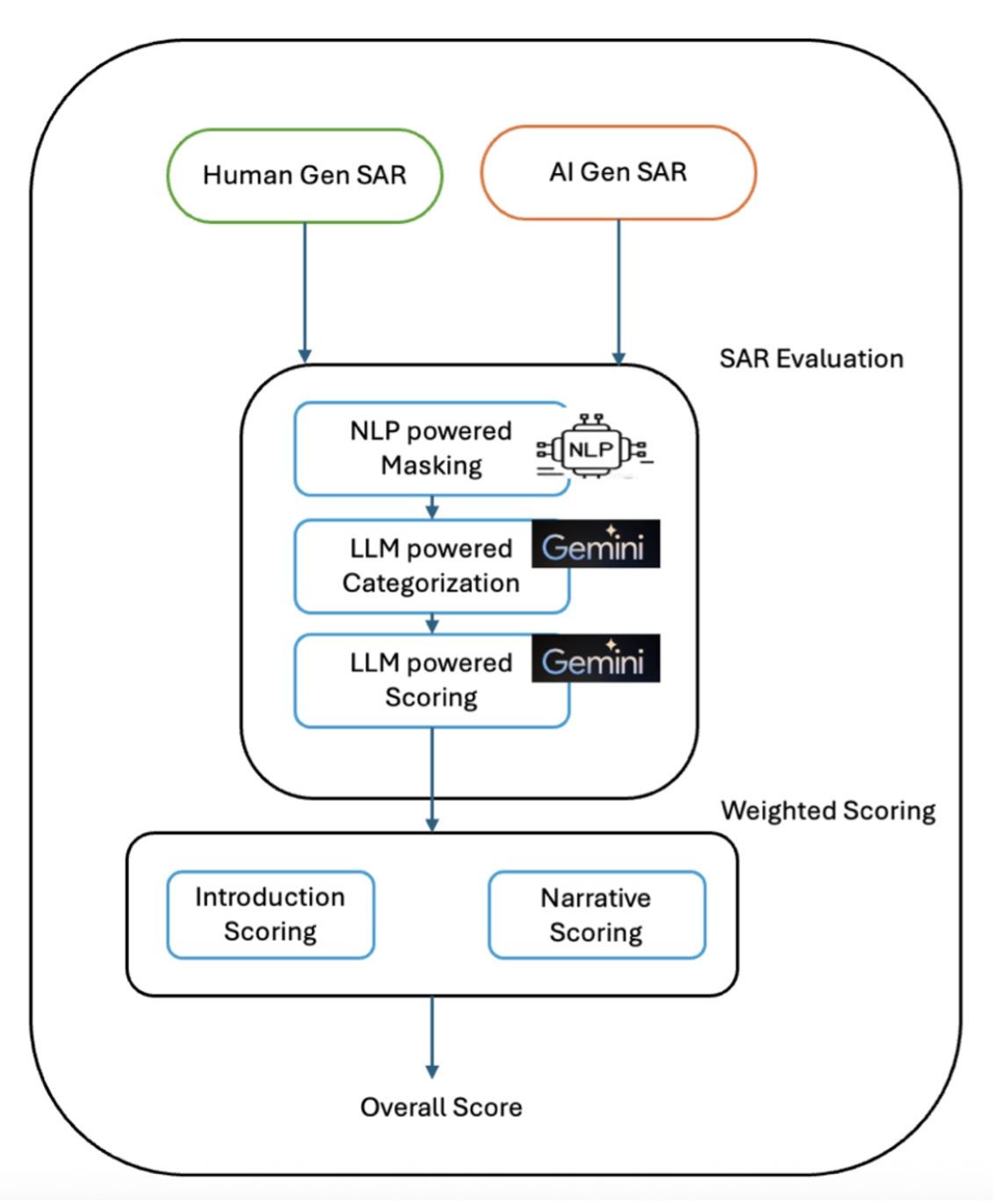}
  \caption{Automated evaluation framework for SAR narrative quality and
regulatory alignment}
  \label{fig:evaluation-framework}
\end{figure}

We incorporate \textbf{Google Gemini 2.5 Pro} as an independent
evaluation LLM, selected for its extended context handling and high
factuality alignment. Gemini is used to perform semantic similarity
scoring, logical coherence analysis, and interpretability assessments
across SAR sections.

\section{Live Evaluation via Compliance Validation Agent (Agent-as-a-Judge)}\label{live-evaluation-via-compliance-validation-agent-agent-as-a-judge}

To ensure the sustained quality and regulatory compliance of Suspicious
Activity Report (SAR) narratives generated by the Co-Investigator AI in
live production environments, we have deployed a dedicated
live-evaluation agent: the \textbf{Compliance Validation Agent},
implementing the \textbf{Agent-as-a-Judge} methodology\hyperlink{ref-5}{[5]}. This
agent autonomously and continuously validates narrative outputs
immediately upon generation, systematically ensuring their accuracy,
coherence, and compliance alignment, and facilitating refinement.

\subsection{Integration and Workflow within Agentic Architecture}\label{integration-and-workflow-within-agentic-architecture}

The Compliance Validation Agent operates seamlessly within the existing
framework, interacting dynamically with multiple components to
comprehensively evaluate live-generated narratives. Its operational
workflow and integrations are detailed as follows:

\begin{itemize}
\item
  \textbf{Coordination with Planning Agent:}
  Upon narrative completion by narrative agent, the Planning Agent
  activates the Compliance Validation Agent, triggering real-time
  narrative assessments based on predefined compliance priorities and
  regulatory risk profiles.
\item
  \textbf{Interaction with Specialized Typology Detection Agents:}
  The Compliance Validation Agent references detailed outputs from
  specialized typology detection agents---including Transaction Fraud
  Detection, Geographic Anomaly Detection, Country Risk Detection, and
  others---to verify the narrative's accuracy and completeness. It
  ensures narratives accurately reflect insights from typology-specific
  risk assessments.
\item
  \textbf{Dynamic Memory Utilization:}
  To conduct precise evaluations, the Compliance Validation Agent
  accesses critical dynamic memory layers:

  \begin{itemize}
  \item
    \textbf{Regulatory Memory:} Retrieves updated AML standards,
    compliance guidelines, and narrative-quality benchmarks to ensure
    regulatory compliance.
  \item
    \textbf{Historical Narrative Memory:} Cross-references previously
    approved narratives, ensuring consistency and continuity of
    narrative structure and quality.
  \item
    \textbf{Typology-Specific Memory:} Accesses historical typological
    insights and past narrative-quality outcomes, verifying the
    consistency and accuracy of risk assessments.
  \end{itemize}
\item
  \textbf{Feedback Agent Interaction:}
  Identified narrative-quality concerns or compliance discrepancies are
  immediately communicated to the Feedback Agent, triggering
  investigator reviews and iterative narrative refinements, thus
  maintaining robust human-in-the-loop oversight.
\end{itemize}

\subsection{Real-Time Evaluation Methodology}\label{real-time-evaluation-methodology}

In live operational environments, the \textbf{Compliance Validation
Agent (Agent-as-a-Judge)} systematically evaluates generated SAR
narratives, ensuring their accuracy, interpretability, and regulatory
compliance. This agent operates in close synchronization with
specialized typology detection agents, performing structured evaluations
through the following rigorous criteria:

\begin{itemize}
\item
  \textbf{Semantic Coherence and Narrative Interpretability:}

  \begin{itemize}
  \item
    Utilizes semantic coherence analyses powered by Large Language
    Models (LLMs) to ensure narratives clearly articulate complex
    compliance scenarios, logical reasoning, and risk interpretations,
    facilitating investigator comprehension and regulatory alignment.
  \end{itemize}
\item
  \textbf{Accuracy Verification with Typology Agent Insights:}

  \begin{itemize}
  \item
    Synchronizes with outputs from specialized typology detection agents
    (e.g., Transaction Fraud Detection Agent, Geographic Anomaly
    Detection Agent, Country Risk Detection Agent) to obtain detailed
    risk indicators, analytical outcomes, and typology-specific data.
  \item
    Systematically cross-validates narrative elements---such as
    identified risks, transaction patterns, subject behaviors, and
    contextual anomalies---against these agent-generated analyses,
    verifying consistency and accuracy.
  \item
    Employs structured rule-based logic to detect any mismatches,
    discrepancies, or omissions between specialized
    agents\textquotesingle{} outputs and corresponding narrative
    descriptions, promptly flagging such inconsistencies for immediate
    investigator attention and remediation.
  \end{itemize}
\item
  \textbf{Transactional and Subject Detail Verification:}

  \begin{itemize}
  \item
    Performs systematic verification of critical narrative
    details---including transaction amounts, date ranges, jurisdictional
    information, and customer identities---against original parsed data
    and typology agents\textquotesingle{} derived insights, ensuring
    precise factual consistency.
  \end{itemize}
\item
  \textbf{Regulatory Compliance Adherence:}

  \begin{itemize}
  \item
    References Regulatory Memory to ensure narratives align with
    evolving AML standards and typology-specific regulatory
    expectations, verifying the completeness of required regulatory
    elements such as external negative-news references and comprehensive
    risk assessments.
  \end{itemize}
\end{itemize}

Through these structured synchronization and cross-validation steps with
specialized typology detection agents, the Compliance Validation Agent
rigorously ensures narrative accuracy, coherence, and regulatory
adherence, thereby significantly enhancing investigator confidence and
operational compliance effectiveness in live environments.

\subsection{Continuous Improvement and Iterative Refinement}\label{continuous-improvement-and-iterative-refinement}

The \textbf{Compliance Validation Agent} enables rapid iterative
refinement by providing immediate narrative-quality feedback, quickly
identifying and addressing inaccuracies. Investigator inputs, driven by
agent-flagged concerns, are dynamically integrated into narrative
generation, continually refining future outputs. Real-time evaluation
results further inform adaptive updates to agent models and dynamic
memory, ensuring sustained narrative accuracy, robust regulatory
compliance, and enhanced investigator trust.

\section{Human-Centered Design and Investigator Collaboration}\label{human-centered-design-and-investigator-collaboration}

Recognizing the inherent interpretive complexities and nuanced
investigative judgment involved in compliance scenarios, the
Co-Investigator AI framework explicitly adopts a human-centered design
philosophy. Rather than implementing a fully automated, "approve-and-go"
approach, the system strategically produces comprehensive initial
narrative drafts, intentionally designed for human refinement and expert
validation.

To facilitate effective collaboration, our framework provides
investigators with a specialized narrative editing interface securely
integrated within protected environments. This interface enables
investigators to easily adjust, enhance, and validate narrative outputs.
Structured feedback loops are embedded directly into the editing
workflow, ensuring investigator inputs are systematically captured and
integrated, thus allowing iterative narrative refinement based on
ongoing human expertise and regulatory insights. This design aligns with
best practices in human-AI collaboration, emphasizing transparency, user
agency, and joint decision-making\hyperlink{ref-32}{[32]}\hyperlink{ref-36}{[36]}. Studies show that
collaborative human-AI workflows improve accountability, trust, and
decision quality---making them especially critical in regulated domains.

\FloatBarrier
\begin{table*}[t]
\centering
\caption{Expert Ratings Across Co-Investigator AI Components}
\begin{tabularx}{\textwidth}{@{}Xcc@{}}
\toprule
\textbf{Component} & \textbf{Avg. Score (1--5)} & \textbf{Effectiveness (\%)} \\
\midrule
Narrative Completeness (Combined intro \& draft) & 3.52 & 70\% \\
Efficiency Gains in Investigation & 3.06 & 61\% \\
Financial Transaction Analysis & 3.50 & 70\% \\
Volume/Velocity Anomaly Detection & 4.00 & 80\% \\
Jurisdictional Risk Assessment & 3.80 & 76\% \\
Communications/Text Pattern Detection & 4.00 & 80\% \\
Location-Based Anomaly Detection & 5.00 & 100\% \\
Account Integrity Monitoring & 4.67 & 93\% \\
Dispute \& Chargeback Pattern Analysis & 4.50 & 90\% \\
\bottomrule
\end{tabularx}
\end{table*}
\FloatBarrier

\section{Empirical Evaluation by Domain Experts}
A comprehensive evaluation was conducted by a team of six domain-expert
investigators from a leading global fintech company, each with
multi-year experience in AML investigations, SAR reporting, and compliance
 operations. They independently assessed each AI‑generated SAR draft, 
 focusing on two key aspects: accuracy of the draft and the potential 
 time savings it offers the investigator. Table 1 provides a summary of the aggregated results.

\textbf{Evaluation Context}: Investigators rated the drafts for
\textbf{accuracy} (how well the AI captured and structured key SAR
elements) and \textbf{efficiency} (time saved using the draft). Scores
were averaged across evaluators and normalized to percentage values for
interpretation.

\textbf{Key Takeaways}

\begin{itemize}
\item
  \textbf{Narrative Completeness (70\%)}: Demonstrates solid coverage of
  core SAR elements. In specific crime typologies, this metric even
  reaches 87\%, requiring only targeted edits during final investigator
  review.
\item
  \textbf{Time Savings (61\%)}: Indicates meaningful efficiency gains by
  reducing manual drafting time, while still enabling hands-on
  oversight.
\item
    \textbf{Module Highlights}: Strong performance in specialized detection modules
    \begin{itemize}
    \item
      \textbf{Location Anomaly Detection (100\%)}: Consistently flagged
      geographic irregularities with perfect accuracy.
    \item
      \textbf{Account Integrity Monitoring (93\%)}: Demonstrated high 
      confidence in detecting compromised or misused accounts.
    \item
      \textbf{Dispute \& Chargeback Analysis (90\%)}: Effectively surfaced
      questionable dispute activity patterns.
    \end{itemize}
\item
  \textbf{Improvement Opportunity Areas}: Financial transaction and
  jurisdictional risk modules show solid baseline performance with
  opportunities to enhance their contribution to investigative insights.
\end{itemize}

This expert-driven assessment confirms the Co-Investigator AI's strong
performance in accuracy and efficiency and highlights how integrating
human-in-the-loop processes can elevate draft precision to meet
regulatory and investigative expectations.



\section{Lessons Learned and Future Work}\label{lessons-learned-and-future-work}

Through the development and deployment of the Co-Investigator AI
framework, we identified several key insights:

\begin{itemize}
\item
  \textbf{Modular and Agentic Approaches:}
  Modular agent architectures, specialized typology detection, and
  structured human-in-the-loop collaboration significantly enhance
  narrative accuracy, compliance adherence, and investigator
  productivity.
\item
  \textbf{Human-AI Collaboration:}
  Explicit integration of human investigator insights strengthens
  narrative quality, regulatory alignment, and trust, highlighting the
  critical role of balanced human-AI interaction.
\item
  \textbf{Real-Time Validation Benefits:}
  Implementing agent-based live evaluation methods effectively ensures
  prompt identification and resolution of compliance issues, minimizing
  operational and regulatory risks.
\item
  \textbf{Enhanced Explainability and Reasoning:}
  Incorporating explicit Chain-of-Thought reasoning methods and
  structured confidence scoring significantly improves system
  transparency, investigator interpretability, and narrative
  accountability.
\end{itemize}

Future directions for this research include:

\begin{itemize}
\item
  \textbf{Expanding Crime Typology Coverage:}
  Extending analytical capabilities to address emerging financial crime
  typologies and evolving regulatory priorities more comprehensively.
\item
  \textbf{Advanced Regulatory Validation:}
  Further developing automated compliance validation methods to
  proactively align with evolving Anti-Money Laundering (AML) standards
  and regulatory guidelines.
\item
  \textbf{Robust Explainability and Auditability:}
  Introducing enhanced explainability frameworks, detailed reasoning
  visualizations, and comprehensive audit trails for each agent to
  strengthen transparency and facilitate regulatory audits.
\item
  \textbf{Adaptive Learning Systems:}
  Developing adaptive mechanisms that dynamically integrate evolving
  regulatory standards, investigator feedback, and reasoning outcomes,
  ensuring continuous improvement, compliance alignment, and system
  reliability.
\end{itemize}

\section{Conclusion}\label{conclusion}

In this work, we presented the \textbf{Co-Investigator AI}, a modular
agentic framework that augments compliance investigations through
automated yet human-guided SAR narrative generation. Leveraging a robust
architecture---featuring the crime typology detection, data privacy via
AI-Privacy Guard, reasoning-enhanced narrative agents, and MCP-enabled
external intelligence---the system delivers accurate and explainable
draft narratives 70\% complete on average, saving investigators
significant time (around 61\%). Our evaluation with domain experts from
a leading global fintech institution confirmed strong performance in key
analytic areas such as geography and account anomaly detection.

By combining domain expertise with Agentic AI and protocol-driven
intelligence access, the Co-Investigator AI improves both the
\textbf{efficiency} and \textbf{integrity} of AML compliance processes.
While the human-in-the-loop remains essential for final validation, our
work demonstrates that AI can substantially reduce the burden of routine
tasks and elevate investigative quality. Future efforts will aim to
enhance robustness, broaden typological coverage, and introduce adaptive
learning mechanisms to further strengthen regulatory compliance and
investigator trust.

\section*{Acknowledgments}
We are grateful to Lisa Davidson, Jeff Maxwell, Oliver Van Zant, Eileen Dinnan, and the broader investigations team for their support in reviewing aspects of the agentic system and for their input, which has been instrumental in shaping its development to the current stage.
\section{References}\label{references}

\begin{enumerate}
\def\labelenumi{\arabic{enumi}.}
\item\hypertarget{ref-1}{ Anthropic. Building effective agents. https://www.anthropic.com/research/building-effective-agents. November, 2024.}
\item\hypertarget{ref-2}{
  Juraj Gottweis, Wei-Hung Weng, Alexander Daryin, Tao Tu, Anil Palepu,
  Petar Sirkovic, Artiom Myaskovsky, Felix Weissenberger, Keran Rong,
  Ryutaro Tanno, et al. Towards an ai co-scientist. arXiv preprint
  arXiv:2502.18864, 2025.}
\item\hypertarget{ref-3}{
  Jason Wei, Xuezhi Wang, Dale Schuurmans, Maarten Bosma, brian ichter,
  Fei Xia, Ed Chi, Quoc V Le, and Denny Zhou. 2022. Chain-of-thought
  prompting elicits reasoning in large language models. In Advances in
  Neural Information Processing Systems, volume 35, pages 24824--24837.
  Curran Associates, Inc.}
\item\hypertarget{ref-4}{
  Han, J., Huang, Y., Liu, S., \& Towey, K. (2020). Artificial
  intelligence for anti-money laundering: a review and extension.
  \emph{Digital Finance}, 2, 181--210.}
\item\hypertarget{ref-5}{
  Zhuge, M., Zhao, C., Ashley, D., Wang, W., Khizbullin, D., Xiong, Y.,
  Liu, Z., Chang, E., Krishnamoorthi, R., Tian, Y., et al.
  Agent-as-a-judge: Evaluate agents with agents. arXiv preprint
  arXiv:2410.10934, 2024.}
\item\hypertarget{ref-6}{
  Amoako, E.K.W., Boateng, V., Ajay, O., Adukpo, T.K., Mensah, N.
  (2025). Exploring the Role of Machine Learning and Deep Learning in
  Anti-Money Laundering (AML) Strategies within U.S. Financial Industry:
  A Systematic Review of Implementation, Effectiveness, and Challenges.
  Finance \& Accounting Research Journal, 7(1).
  \url{https://doi.org/10.51594/farj.v7i1.1808}}
\item\hypertarget{ref-7}{
  Xu, Ziwei, Sanjay Jain, \& Mohan Kankanhalli. 2024. Hallucination Is
  Inevitable: An Innate Limitation of Large Language Models.
  doi:10.48550/arXiv.2401.11817}
\item\hypertarget{ref-8}{
  Ji, Z.; Yu, T.; Xu, Y.; Lee, N.; Ishii, E.; Fung, P. Towards
  mitigating LLM hallucination via self reflection. In Proceedings of
  the 2023 Conference on Empirical Methods in Natural Language
  Processing, Singapore, 6--10 December 2023.}
\item\hypertarget{ref-9}{
  Z. Xi, W. Chen, X. Guo, W. He, Y. Ding, B. Hong, M. Zhang, J. Wang, S.
  Jin, E. Zhou, et al., The rise and potential of large language model
  based agents: A survey, arXiv preprint arXiv:2309.07864 (2023)}
\item\hypertarget{ref-10}{
  Yuheng Cheng, Ceyao Zhang, Zhengwen Zhang, Xiangrui Meng, Sirui Hong,
  Wenhao Li, Zihao Wang, Zekai Wang, Feng Yin, Junhua Zhao, et al.
  Exploring large language model based intelligent agents: Definitions,
  methods, and prospects. arXiv preprint arXiv:2401.03428, 2024.}
\item\hypertarget{ref-11}{
  S. Kusal, S. Patil, J. Choudrie, K. Kotecha, S. Mishra, and A.
  Abraham, ``Ai-based conversational agents: A scoping review from
  technologies to future directions,'' IEEE Access, 2022.}
\item\hypertarget{ref-12}{
  C. Ma, J. Li, K. Wei, B. Liu, M. Ding, L. Yuan, Z. Han, and H. V.
  Poor, ``Trusted ai in multiagent systems: An overview of privacy and
  security for distributed learning,'' Proceedings of the IEEE, vol.
  111, no. 9, pp. 1097--1132, 2023.}
\item\hypertarget{ref-13}{
  Wu J, Huang Z, Hu Z, Lv C. Toward human-in-the-loop AI: Enhancing deep
  reinforcement learning via real-time human guidance for autonomous
  driving. Engineering. 2023; 21:75--91.}
\item\hypertarget{ref-14}{
  Shavit, S. Agarwal, M. Brundage, S. Adler, C. O'Keefe, R. Campbell, T.
  Lee, P. Mishkin, T. Eloundou, A. Hickey, K. Slama, L. Ahmad, P.
  McMillan, A. Beutel, A. Passos, and D. G. Robinson. Practices
  governing agentic AI systems. https://cdn.openai.com/pap
  ers/practices-for-governing-agentic-ai-systems.pdf, 2023. Accessed:
  2023-01-04.}
\item\hypertarget{ref-15}{
  Chan, R. Salganik, A. Markelius, C. Pang, N. Rajkumar, D.
  Krasheninnikov, L. Langosco, Z. He, Y. Duan, M. Carroll, M. Lin, A.
  Mayhew, K. Collins, M. Molamohammadi, J. Burden, W. Zhao, S. Rismani,
  K. Voudouris, U. Bhatt, A. Weller, D. Krueger, and T. Maharaj. Harms
  from increasingly agentic algorithmic systems. In 2023 ACM Conference
  on Fairness, Accountability, and Transparency, FAccT '23. ACM, June
  2023b. doi: 10.1145/3593013.3594033. URL
  \url{http://dx.doi.org/10.1145/3593013.3594033}.}
\item\hypertarget{ref-16}{
  Y. Liu, M. Ott, N. Goyal, J. Du, M. Joshi, D. Chen, O. Levy, M. Lewis,
  L. Zettlemoyer, V. Stoyanov, Roberta: A robustly optimized bert
  pretraining approach, arXiv preprint arXiv:1907.11692 (2019).}
\item\hypertarget{ref-17}{
  Shuai Zheng, Sadeep Jayasumana, Bernardino RomeraParedes, Vibhav
  Vineet, Zhizhong Su, Dalong Du, Chang Huang, and Philip H. S. Torr.
  Conditional random fields as recurrent neural networks. In ICCV, 2015.
  2}
\item\hypertarget{ref-18}{
  \href{https://www.spiedigitallibrary.org/profile/Dinesh.Verma-15863}{Dinesh
  C. Verma} and \ul{R. Ratnaparkhi} "Guardrails for safe implementations
  of AI-based services", Proc. SPIE 13476, Assurance and Security for
  AI-enabled Systems 2025, 134760I (28 May 2025);
  \url{https://doi.org/10.1117/12.3051891}}
\item\hypertarget{ref-19}{
  Y. Dong, R. Mu, G. Jin, Y. Qi, J. Hu, X. Zhao, J. Meng, W. Ruan, and
  X. Huang, ``Building guardrails for large language models,'' arXiv
  preprint arXiv:2402.01822, 2024.}
\item\hypertarget{ref-20}{
  Z. Yuan, Z. Xiong, Y. Zeng, N. Yu, R. Jia, D. Song, and B. Li,
  ``Rigorllm: resilient guardrails for large language models against
  undesired content,'' in Proceedings of the 41st International
  Conference on Machine Learning, 2024, pp. 57 953--57 965.}
\item\hypertarget{ref-21}{
  Z. Wang, F. Yang, L. Wang, P. Zhao, H. Wang, L. Chen, Q. Lin, and
  K.-F. Wong, ``SELF-GUARD: Empower the LLM to safeguard itself,'' in
  Proceedings of the 2024 Conference of the North American Chapter of
  the Association for Computational Linguistics, 2024, pp. 1648--1668.}
\item\hypertarget{ref-22}{
  S. Bakhtiari, Z. Nasiri, and J. Vahidi, `\,`Credit card fraud
  detection using ensemble data mining methods,'\,' Multimedia Tools
  Appl., vol. 82, no. 19, pp. 29057--29075, Aug. 2023.}
\item\hypertarget{ref-23}{
  Biswas A, Talukdar W. Building Agentic AI Systems: Create intelligent,
  autonomous AI agents that can reason, plan, and adapt. 1st ed. Packt
  Publishing 2025.}
\item\hypertarget{ref-24}{
  X. Hou, Y. Zhao, S. Wang, and H. Wang, ``Model context protocol (mcp):
  Landscape, security threats, and future research directions,'' arXiv
  preprint arXiv:2503.23278, 2025.}
\item\hypertarget{ref-25}{
  Satyadhar Joshi, Review of Gen AI Models for Financial Risk
  Management, International Journal of Scientific Research in Computer
  Science, Engineering and Information Technology, vol. 11, no. 1, pp.
  709--723, Jan. 2025, doi: 10.32628/CSEIT2511114.}
\item\hypertarget{ref-26}{
  Satyadhar Joshi, ``Leveraging prompt engineering to enhance financial
  market integrity and risk management,'' World Journal of Advanced
  Research and Reviews, vol. 25, no. 1, pp. 1775--1785, 2025.}
\item\hypertarget{ref-27}{
  T. Masterman, S. Besen, M. Sawtell and A. Chao, arXiv, 2024, preprint,
  arXiv:2404.11584, DOI: 10.48550/ arXiv.2404.11584.}
\item\hypertarget{ref-28}{
  S. Yao, D. Yu, J. Zhao, I. Shafran, T. L. Griffiths, Y. Cao and K.
  Narasimhan, arXiv, 2023, preprint, arXiv:2305.10601, DOI:
  10.48550/arXiv.2305.10601.}
\item\hypertarget{ref-29}{
  Z. Gou, Z. Shao, Y. Gong, Y. Shen, Y. Yang, N. Duan and W. Chen,
  arXiv, 2024, preprint, arXiv:2305.11738, DOI:
  10.48550/arXiv.2305.11738.}
\item\hypertarget{ref-30}{
  Edward Yeo, Yuxuan Tong, Morry Niu, Graham Neubig, and Xiang Yue.
  Demystifying long chain-of-thought reasoning in llms. arXiv preprint
  arXiv:2502.03373, 2025.}
\item\hypertarget{ref-31}{
  Shunyu Yao, Noah Shinn, Pedram Razavi, and Karthik Narasimhan.
  \$\textbackslash tau\$-bench: A Benchmark for Tool-Agent-User
  Interaction in Real-World Domains, June 2024. URL
  http://arxiv.org/abs/2406.12045. arXiv:2406.12045 {[}cs{]}.}
\item\hypertarget{ref-32}{
  Matéo Mahaut, Laura Aina, Paula Czarnowska, Momchil Hardalov, Thomas
  Müller, and Lluís Màrquez. 2024. Factual Conidence of LLMs: on
  Reliability and Robustness of Current Estimators. arXiv preprint
  arXiv:2406.13415 (2024).}
\item\hypertarget{ref-33}{
  Madaan A, Tandon N, Gupta P, Hallinan S, Gao L, Wiegreffe S, Alon U,
  Dziri N, Prabhumoye S, Yang Y, Gupta S, Majumder B P, Hermann K,
  Welleck S, Yazdanbakhsh A, Clark P. Self-refine: iterative refinement
  with self-feedback. Advances in Neural Information Processing Systems,
  2024, 36.}
\item\hypertarget{ref-34}{
  Jiazheng Kang, Mingming Ji, Zhe Zhao, Ting Bai., et al. (2024).
  \emph{MemoryOS: A Memory Operating System for Building Personal LLM
  Agents}. arXiv preprint arXiv:2506.06326.
  \url{https://arxiv.org/abs/2506.06326}}
\item\hypertarget{ref-35}{
  Wujiang Xu, Kai Mei, Hang Gao, Juntao Tan, Zujie Liang, Yongfeng
  Zhang., et al. (2024). \emph{A-MEM: An Agentic Structured Memory
  Network for LLM Agents}. arXiv preprint arXiv:2502.12110.
  \url{https://arxiv.org/abs/2502.12110}}
\item\hypertarget{ref-36}{
  Amershi, S., Weld, D., Vorvoreanu, M., et al. (2019). \emph{Guidelines
  for Human-AI Interaction}. Proceedings of the 2019 CHI Conference on
  Human Factors in Computing Systems.}
\item\hypertarget{ref-37}{
  Patrick Lewis, Ethan Perez, Aleksandra Piktus, Fabio Petroni, Vladimir
  Karpukhin, Naman Goyal, Heinrich Küttler, Mike Lewis, Wen-tau Yih, Tim
  Rocktäschel, et al. Retrieval-augmented generation for
  knowledge-intensive nlp tasks. Advances in Neural Information
  Processing Systems, 33:9459--9474, 2020.}
\end{enumerate}

\end{document}